\pdfoutput=1

\documentclass[11pt]{article}

\usepackage[final]{acl}

\usepackage{times}
\usepackage{latexsym}

\usepackage[T1]{fontenc}

\usepackage[utf8]{inputenc}

\usepackage{microtype}

\usepackage{inconsolata}
\usepackage{graphicx}
\usepackage{xspace,mfirstuc,tabulary}
\usepackage[font=small,skip=0pt]{caption}
\usepackage{subcaption}
\usepackage{float}
\usepackage{makecell}

\title{eagerlearners at SemEval2024 Task 5: The Legal Argument Reasoning Task in Civil Procedure}

\author{Hoorieh Sabzevari, Mohammadmostafa Rostamkhani, Sauleh Eetemadi \\
        Iran University of Science and Technology \\
        \small \texttt{h\_sabzevari@elec.iust.ac.ir, mo\_rostamkhani97@comp.iust.ac.ir, sauleh@iust.ac.ir }
        }

\begin{document}
\maketitle
\begin{abstract}
This study investigates the performance of the zero-shot method in classifying data using three large language models, alongside two models with large input token sizes and the two pre-trained models on legal data. Our main dataset comes from the domain of U.S. civil procedure. It includes summaries of legal cases, specific questions, potential answers, and detailed explanations for why each solution is relevant, all sourced from a book aimed at law students. By comparing different methods, we aimed to understand how effectively they handle the complexities found in legal datasets. Our findings show how well the zero-shot method of large language models can understand complicated data. We achieved our highest \begin{math}F_{1}\end{math} score of 64\% in these experiments.
\end{abstract}

\section{Introduction}
Becoming skilled at presenting a legal case is essential for aspiring lawyers. It requires understanding not only the relevant legal areas but also using advanced reasoning tactics like making analogies and spotting hidden contradictions.  \cite{chalkidis-etal-2022-lexglue}. Despite efforts to set standards for modern NLP models in legal language understanding, there still aren't complex tasks focusing on argumentation in legal matters.\cite{Bongard.et.al.2022.NLLP}

The dataset utilized in this study was gathered from \emph{The Glannon Guide To Civil Procedure} by Joseph Glannon \cite{glannon2018glannon} in English. Each sample within the dataset comprises a question, a solution, and an introduction elaborating on the provided solution. The objective is to identify whether the given answer, derived from the introduction text, accurately addresses the question. 

This paper explores various approaches to address the challenge of handling lengthy and intricate data, which can be challenging for human comprehension. Initially, we evaluated two models—Longformer \cite{beltagy2020longformer} and Big Bird \cite{zaheer2021big} —known for their effectiveness in classifying data with large input tokens. Subsequently, we assessed the performance of two pre-trained models, Legal-RoBERTa \cite{chalkidis-garneau-etal-2023-lexlms} and Legal-XLM-RoBERTa \cite{Niklaus2023MultiLegalPileA6} for legal data using the original code. Finally, we compared the performance of three large language models —GPT 3.5, Gemini, and Copilot— using the zero-shot method on the test dataset.

We recognized the significant impact of leveraging the capabilities and extensive capacity of large language models on analyzing data, especially those focusing on specific topics or lengthy content. Looking ahead, our goal is to improve prompts further to achieve superior results not only for this task but also for similar works. Further details regarding the implementation can be found in \href{https://github.com/lhoorie/SemEval2024-Task5.git}{this GitHub repository}.

\section{Background}
\subsection{Task Setup}
As previously mentioned, the original dataset for this task is sourced from \emph{The Glannon Guide To Civil Procedure}. Each sample within the dataset comprises the following components:

1. Question
2. Answer
3. Label
4. Analysis
5. Complete Analysis
6. Explanation

It's noteworthy that "Analysis" and "Complete Analysis" were absent in the test data. The data split involves allocating the initial 80\% of questions from each chapter to the training set, the subsequent 10\% to the validation set, and the final 10\%—typically more challenging questions—to the test set. The final dataset consists of 848 entries.

In this task, the inputs to the model consist of the "Question," "Explanation," and "Answer" values, while the output of the model is represented by the "Label." If the model determines that the answer provided for the question aligns with the explanation, the output label will be 1; otherwise, it will be 0. The objective of this task is to assess the model's reasoning capabilities, particularly its ability to analyze legal issues effectively.

\subsection{Related Work}
\subsubsection{Pre-trained Legal Language Models}
Numerous studies have been conducted in the realm of legal issues. Given the challenges posed by comprehending lengthy texts within this domain using existing models, pre-trained language models have been tailored to address this need. One such model is the Legal-BERT model \cite{chalkidis2020legalbert}, which is also employed in this paper. This study introduces a specialized model aimed at facilitating NLP-based legal research by fine-tuning the original BERT model for legal applications. \cite{li2023sailer} introduces SAILER, a novel pre-trained linguistic model with a unique architecture designed for legal case retrieval. Furthermore, \cite{10255647} provides a comprehensive survey of existing Legal Judgment Prediction (LJP) tasks, datasets, and models within the legal domain, encompassing an overview of 8 pre-trained models across 4 languages as part of the LJP. Moreover, an end-to-end methodology is introduced by \cite{louis2023interpretable} for generating long-form answers to statutory law questions, addressing limitations in existing Legal Question Answering (LQA) approaches.
\subsubsection{Domain-Specific LLMs in Law}
A Large Language Model (LLM), such as ChatGPT, is remarkable for its ability to handle general-purpose language generation and a variety of other NLP tasks.
Domain-specific LLMs are versatile models optimized to excel at specific tasks defined by organizational standards. They further empower lawyers to expand their understanding and explore specialized legal domains. For instance, \cite{colombo2024saullm7b} is the first LLM designed explicitly for legal text comprehension and generation with 7 billion parameters. To empower the legal field, \cite{cui2023chatlaw} presents ChatLaw, an open-source legal LLM built with a high-quality, domain-specific fine-tuning dataset. The focus of \cite{savelka2023explaining} is evaluating GPT-4's effectiveness in generating explanations for legal terms – specifically, whether they are accurate, clear, and relevant to the surrounding legislation.

\section{System Overview}
\subsection{Preprocessing Data}

Our dataset comprises 666 samples for the training set, 84 samples for the validation set, and 98 samples for the test set. Initially, we excluded the columns "Analysis" and "Complete Analysis" from both the training and validation datasets as they were absent in the test data. Afterward, we analyzed to determine the distribution of class labels 0 and 1, revealing a notable class imbalance, with the number of instances belonging to class 0 nearly three times higher than those of class 1.

To address this issue, various approaches can be employed. In this study, we opted to mitigate the class imbalance using the focal loss function as our loss function. Our investigation demonstrates the efficacy of focal loss in rectifying class imbalance, enhancing the performance of classes with limited training samples, offering adaptability in adjusting the learning process, and attenuating the impact of noisy data.

\subsection{Model}
\subsubsection{Pre-trained Models}
In this study, we tackled the challenge of dealing with long sets of data. To overcome this, we looked into using two models designed to handle large inputs: the Longformer and Big Bird models. These models can handle up to 4096 tokens, which is much more than the BERT model. We also used two pre-trained models specifically trained for legal data: Legal-RoBERTa and Legal-XLM-RoBERTa. These models, like Legal-BERT \cite{chalkidis2020legalbert}, were trained on various legal documents and cases.

Our aim is straightforward: to compare the performance of these models and understand how using pre-trained models with larger input sizes impacts their effectiveness. Through this analysis, we aim to gain insights into the optimal approaches for managing complex legal datasets.

\begin{figure}[H]
\resizebox{\linewidth}{!}{
     \includegraphics[scale=0.25]{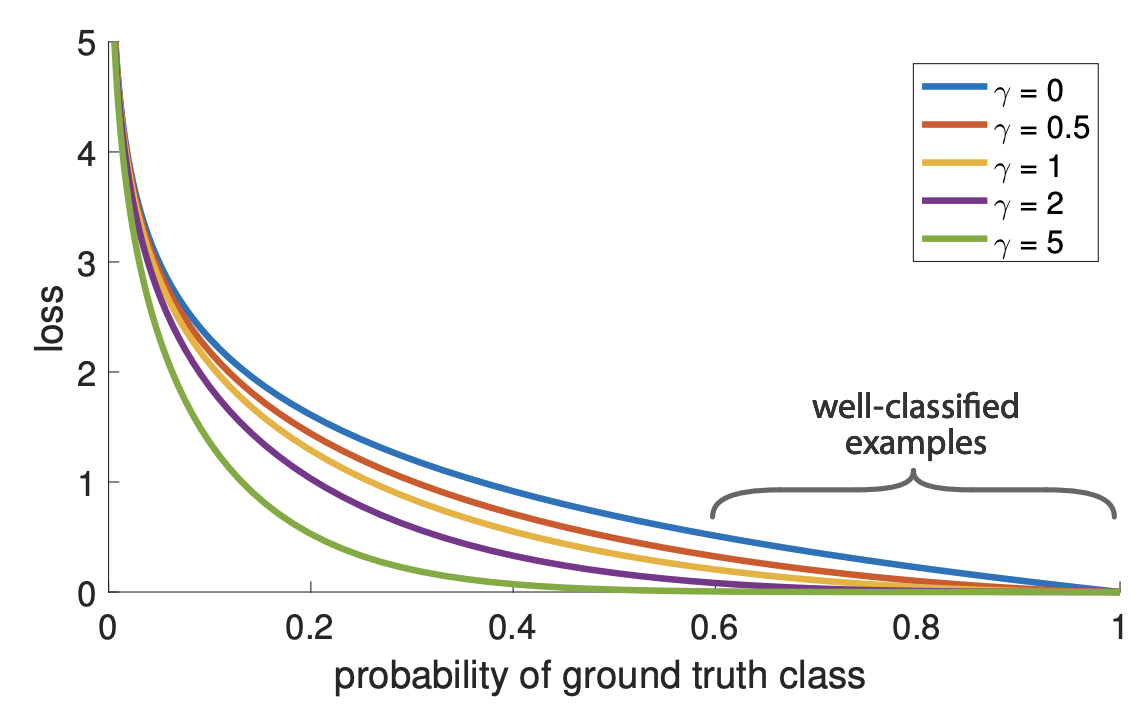}
     }
     \caption{Comparison between cross entropy and focal loss}
    \label{fig:fig-1}
\end{figure}

To address the issue of unbalanced data, we implemented the focal loss function, a method that has shown promising outcomes in previous research \cite{lin2018focal} \cite{wang-etal-2022-calibrating}. The Focal Loss function formally incorporates a factor of \((1-p_{t})^{\gamma}\) into the standard cross-entropy criterion. This adjustment diminishes the relative loss for accurately classified examples \begin{math}(p_{t} > 0.5)\end{math}, thereby intensifying the focus on challenging instances that are misclassified. 
\begin{equation}
    CE(p_{t})= -log(p_{t})
\end{equation}
\begin{equation}
    FL(p_{t})= - (1-p_{t})^{\gamma}log(p_{t})
\end{equation}
There is a tunable focusing parameter \begin{math}\gamma \geq 0 \end{math}.
Figure \ref{fig:fig-1} illustrates the varied impact of this parameter across a range of values.

\subsubsection{Large Language Models}
In addition, we leveraged three popular large language models to assess their performance within the given task: OpenAI's GPT 3.5 model, Google's Gemini model, and Bing's Copilot model. Later, we formulated the task in the form of prompts, refined them through prompt engineering techniques, and presented them to the large language models. Our objective was to extract the desired answer (0 or 1) from the models' responses, thereby assessing their performance and capabilities in handling the task. We conducted extensive tests on numerous prompts to identify the most effective ones. Employing prompt engineering methodologies alongside large language models, we refined these prompts to enhance their performance and effectiveness in achieving our objectives.

\section{Experimental setup}
We utilized a dataset of 848 data points, dividing it into 80\% for the training dataset, 10\% for the validation dataset, and 10\% for the test dataset. Our experimental approach involved exploring three main inquiries:
\begin{enumerate}
\item{Exploration of Longformer and Big Bird models, tailored to effectively process lengthy data inputs.}
\item{Utilization of pre-trained models for legal contexts, including Legal-RoBERTa and Legal-XLM-RoBERTa, to ascertain their efficacy in legal text analysis.}
\item{Implementation of the Zero-shot methodology with prompt feeding across three distinct models: GPT 3.5\footnote{\url{https://chat.openai.com/}}, Gemini\footnote{\url{https://gemini.google.com/app}}, and Copilot\footnote{\url{https://www.bing.com/chat}}, aimed at exploring their adaptability and performance across diverse tasks.}
\end{enumerate}

In the initial phase, we employed the AdamW optimizer function with a learning rate set at 5e-5 for both models, conducting training over 3 epochs.
The specific values chosen for each hyperparameter are listed in Table \ref{tab:table-3}.
\begin{table}
    \centering
    \begin{tabular}{c c } 
     \hline
        \textbf{Hyperparameter} & \textbf{Value}	\\[0.5ex]      
     \hline
        \textbf{Optimizer} &  AdamW	\\
        \textbf{Learning Rate}	& 5e-5 \\
        \textbf{Epochs} & 3	\\
        \textbf{Batch Size} & 1 \\
        \textbf{Loss Function} & Focal Loss \\
     \hline
    \end{tabular}
    
    \vspace*{4mm}
    \caption{Hyperparameter values}
    \label{tab:table-3}
\end{table}

Then, in the second phase, we adapted the original code from the article with the necessary modifications.\footnote{\url{https://github.com/trusthlt/legal-argument-reasoning-task}} The original code featured the utilization of Sliding Window Simple (SWS) and Sliding Window Complex (SWC) methods with the Legal-BERT model. We made adjustments to certain sections of the code to ensure compatibility with the test input data. Throughout our evaluation, we utilized the macro \begin{math}
    F_{1}
\end{math} score as the primary evaluation metric.

\begin{table}
    \centering
    \resizebox{\linewidth}{!}{
    \begin{tabular}{c c c} 
     \hline
        \textbf{Model} & \textbf{Accuracy}	& $\mathbf{F_1}$	\\[0.5ex]      
     \hline
    \textbf{Longformer}	& \textbf{0.79}	& 0.44	\\
    \textbf{Big Bird}	& \textbf{0.79}	& 0.44	\\
    \hline
    \textbf{Legal-BERT}	& 0.75	& \textbf{0.58}	\\
    \textbf{Legal-RoBERTa}	& \textbf{0.79}	& 0.44	\\
    \textbf{Legal-XLM-RoBERTa}	& 0.78	& 0.43	\\
     \hline
    \end{tabular}
    }
    \vspace*{4mm}
    \caption{Accuracy and macro F1-score of first and second parts' models on the validation set}
    \label{tab:table-1}
\end{table}

In the final phase of our experiments, we used the API keys provided by OpenAI and Google. Unfortunately, despite our efforts, we encountered challenges locating the official API for Bing. Given its superior accuracy compared to other models, we decided to manually record the results of the test dataset. Throughout this phase, we employed a variety of prompts and iteratively refined them. Our investigation revealed that incorporating terms such as "step by step" or asking for explanations of the inference steps to achieve the desired outcome positively impacted the model's performance.
Furthermore, we encountered limitations in prompt completeness due to the token size constraints inherent in the input models. For instance, Bing's Copilot supported a maximum input size of 4000 characters, which proved insufficient for processing our long samples. Here is an example of an input prompt:\\

\begin{table}
    \centering
    \begin{tabular}{c c c} 
     \hline
        \textbf{Model} & \textbf{Accuracy}	& $\mathbf{F_1}$	\\[0.5ex]      
     \hline
        \textbf{GPT 3.5} & \textbf{0.69}	& 0.59	\\
        \textbf{Gemini}	& 0.44 & 0.44 \\
        \textbf{Copilot} & 0.67	& \textbf{0.64}	\\
     \hline
    \end{tabular}
    
    \vspace*{4mm}
    \caption{Macro F1-score of large language models on the test set}
    \label{tab:table-2}
\end{table}

\fbox{\begin{minipage}{18em}
I will provide a question, an answer, and an explanation. 
Your task is to determine if the answer is correct based on the explanation provided. 
After reading the explanation, please respond with 'yes' if the answer is correct, 
or 'no' if it is incorrect. \\ \\
\textbf{Question:} \{question\} \\
\textbf{Proposed Answer:} \{answer\} \\
\textbf{Explanation:} \{explanation\} \\ \\
Is the proposed answer correct based on the explanation? (yes or no) \\
Please provide your detailed reason for your choice.\\
Then, reevaluate and check whether the selected answer is logical or not. \\
Please use the following format: \\
\textbf{<selected\_answer>:} yes/no \\
\textbf{<reason>:} your reason for the initial choice \\
\textbf{<reason for logical check>:} your reason for reevaluation 
\end{minipage}}

\section{Results}
After completing the aforementioned three phases, our investigation revealed that despite fine-tuning, existing models struggled to effectively analyze lengthy data within challenging legal contexts, encountering training process issues and yielding suboptimal outputs. Moreover, the most promising result emerged from fine-tuning the Legal-BERT model, serving as the baseline. While we continue to analyze the underlying reasons for this outcome, initial observations suggest that the learning challenge may be linked to the specific characteristics of the dataset employed. Table \ref{tab:table-1} presents the performance metrics of the models from the initial two phases, evaluated on the validation dataset.

When it comes to evaluating the results of the zero-shot method, we identified its considerable potential and Bing's Copilot model emerged as the top performer, surpassing expectations. Following suit, the GPT 3.5 model presented moderate performance, while the Gemini model fell short of expected levels. The success of the Copilot model lies in its ability to address previous challenges associated with GPT models by leveraging real-time information accessible through the internet. Table \ref{tab:table-2} presents the results achieved from employing this method on the test dataset, representing the unofficial results submitted during the post-evaluation phase.

\section{Conclusion}

In this paper, we present methods designed for classifying lengthy legal cases. We divided our exploration into three main parts:

Firstly, we looked into models with large input token sizes such as Longformer and Big Bird. Secondly, we examined pre-trained models specifically fine-tuned for legal data, such as Legal-RoBERTa and Legal-XLM-RoBERTa. Lastly, we tested the zero-shot method across three major language models.

Among these methods, we found that the zero-shot technique and Bing's Copilot model showed the most promising performance. As for future works, we can explore techniques like data summarization, collaborative approaches such as the round table technique, trying various hyperparameters, and refine prompts to further enhance model performance. These efforts have the potential to advance the effectiveness of classification tasks in legal contexts.

As a future work, it would be valuable to explore additional large language models. These models offer extensive capabilities, especially in summarizing lengthy datasets, which could help evaluate various models' performance. However, it is necessary to note that during summarization, some important details might be overlooked.

Another avenue for future research involves testing the effectiveness of a multi-model approach. \cite{chen2023reconcile} This method entails bringing together different large language model agents in a round table conference format. This setup encourages diverse perspectives and discussions to foster consensus. By adopting this approach, researchers can tap into the combined intelligence of multiple models, potentially enriching analysis across various tasks and domains.
In this competition, our team achieved the 17th rank out of 21 groups. Our only submission during the evaluation phase utilized the basic prompt and the GPT model. However, significant improvements were made during the post-evaluation phase, resulting in a much higher level of accuracy.


\bibliography{acl_latex}




\end{document}